\documentclass[lettersize,journal]{IEEEtran}
\usepackage{amsmath,amsfonts}
\usepackage{algorithmic}
\usepackage{algorithm}
\usepackage{array}
\usepackage[caption=false,font=normalsize,labelfont=sf,textfont=rm]{subfig}
\usepackage{textcomp}
\usepackage{stfloats}
\usepackage{url}
\usepackage{verbatim}
\usepackage{graphicx}
\usepackage{cite}
\usepackage{eccvabbrv}
\usepackage{bm}
\usepackage{multirow}
\usepackage{lipsum}  

\DeclareSubrefFormat{myparens}{#1 (#2)}
\makeatletter
\let\NAT@parse\undefined
\makeatother
\usepackage{hyperref}  
\usepackage{cleveref}
\usepackage{booktabs}
\usepackage{color}

\newcommand{\lzh}[1]{{\color{black}#1}}

\begin{document}

\title{SceneBooth: Diffusion-based Framework for Subject-preserved Text-to-Image Generation}

\author{Shang Chai, Zihang Lin, Min Zhou, Xubin Li, Liansheng Zhuang, Houqiang Li
\thanks{Manuscript received April 19, 2021; revised August 16, 2021. This work was supported by the National Natural Science Foundation of China under Grant No.U20B2070 and No.61976199. Shang~Chai, Liansheng~Zhuang and Houqiang Li are with the University of Science and Technology of China, Hefei 230000, China (e-mail: chaishang@mail.ustc.edu.cn; lszhuang@ustc.edu.cn; lihq@ustc.edu.cn). Zihang~Lin, Min~Zhou and Xubin~Li are with the Alibaba Group, Beijing 100102, China (e-mail: \{ linzihang.lzh, yunqi.zm, lxb204722\}@alibaba-inc.com).}
\thanks{This work was done during an internship at Alibaba Group. Corresponding author: Liansheng Zhuang.}
}

\markboth{Journal of \LaTeX\ Class Files,~Vol.~14, No.~8, August~2021}%
{Shell \MakeLowercase{\textit{et al.}}: A Sample Article Using IEEEtran.cls for IEEE Journals}

\maketitle

\begin{abstract}
Due to the demand for personalizing image generation, subject-driven text-to-image generation method, which creates novel renditions of an input subject based on text prompts, has received growing research interest. 
Existing methods often learn subject representation and incorporate it into the prompt embedding to guide image generation, but they struggle with preserving subject fidelity.
To solve this issue, this paper approaches a novel framework named SceneBooth for subject-preserved text-to-image generation, which consumes inputs of a subject image, object phrases and text prompts. Instead of learning the subject representation and generating a subject, our SceneBooth fixes the given subject image and generates its background image guided by the text prompts.  To this end, our SceneBooth introduces two key components, \emph{i.e.}, a multimodal layout generation module and a background painting module. 
The former determines the position and scale of the subject by generating appropriate scene layouts that align with text captions, object phrases, and subject visual information.
The latter integrates two adapters (ControlNet and Gated Self-Attention) into the latent diffusion model to generate a background that harmonizes with the subject guided by scene layouts and text descriptions.  
In this manner, our SceneBooth ensures accurate preservation of the subject's appearance in the output.
Quantitative and qualitative experimental results demonstrate that SceneBooth significantly outperforms baseline methods in terms of subject preservation, image harmonization and overall quality.
\end{abstract}

\begin{IEEEkeywords}
Image Generation, Layout Generation, Text-to-Image.
\end{IEEEkeywords}

\section{Introduction}
\begin{figure}[t!]
\captionsetup[subfloat]{font={scriptsize,rm},labelfont=scriptsize}
\centering
\subfloat[Subject images]{\includegraphics[width=0.3\linewidth]{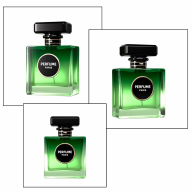}}%
\label{fig:d_a}
\subfloat[Dreambooth]{\includegraphics[width=0.33\linewidth]{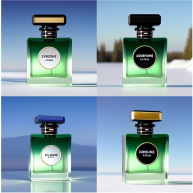}%
\label{fig:d_b}}
\hfil
\subfloat[SceneBooth]{\includegraphics[width=0.33\linewidth]{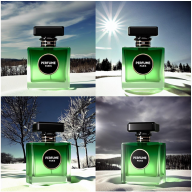}%
\label{fig:d_c}}
\caption{Results generated by subject-driven and subject-perseved text-to-image methods. Text prompt is ``A perfume is placed in the snow." (a) Subject images. (b) Results generated by subject-driven method Dreambooth~\cite{Ruiz2022DreamBoothFT}. There is noticeable distortion in the color and appearance of the text ``PERFUME PARIS". (c) Results generated by our subject-preserved method SceneBooth. The appearance of the perfume is well preserved.}
\label{fig:describe}
\end{figure}

\IEEEPARstart{T}{ext-to-image} generation with user-specific subjects facilitates a wide range of potential applications. For example, in advertising scenarios, advertisers can showcase their products in a visually engaging virtual image to attract potential customers. Likewise, individuals may want to replace the background of their selfies with famous landmarks, creating eye-catching images. Recently, with the impressive progress in large-scale text-to-image models, this area has attracted increasing attention~\cite{kumari2023multi, Ruiz2022DreamBoothFT, DBLP:conf/cvpr/YangGZZCSCW23, DBLP:conf/iclr/GalAAPBCC23, li2023blip}. However, most current methods often fail to accurately preserve the given subject's appearance and thus are not applicable in high-fidelity demanding scenarios. To tackle this problem, we introduce a novel task called subject-preserved text-to-image generation, which ensures the precise preservation of subjects by nature. In contrast to subject-driven image generation~\cite{Ruiz2022DreamBoothFT, chen2023disenbooth, kumari2023multi, DBLP:conf/iclr/GalAAPBCC23, li2023blip,chen2023anydoor}, our proposed task retains the original subject image as the foreground and generates a harmonious background for it with the guidance of a scene caption, the subject image, and object phrases which describe each object in the scene. This task setting brings new challenges, since we need to consider many factors, such as the size and position of the subjects, their semantics, and their relationships to the scene. Fig.~\ref{fig:describe} shows examples generated by subject-driven and subject-preserved text-to-image method.

Generating images with a specific subject has been widely studied. Some early methods~\cite{DBLP:journals/ijcv/AzadiPED20, DBLP:conf/mm/LuLC0023, DBLP:conf/eccv/ZhouLNZ22} first retrieve suitable subject-background pairs and then blend and harmonize them. This process requires a considerable amount of candidate images and often fails to produce harmonious results in terms of geometry, semantics, and lighting~\cite{DBLP:journals/ijcv/LiZMT22}. Recently, subject-driven text-to-image methods~\cite{Ruiz2022DreamBoothFT, chen2023disenbooth, kumari2023multi, DBLP:conf/iclr/GalAAPBCC23}, like DreamBooth~\cite{Ruiz2022DreamBoothFT}, generate images of a given subject by finetuning a pre-trained text-to-image model on multiple subject images. Though having shown impressive success in generating high-quality images, those methods face a trade-off between subject fidelity and background diversity, due to the potential overfitting and language drift~\cite{lee2019countering, lu2020countering, Ruiz2022DreamBoothFT} problem. Moreover, subject distortion can almost always be observed in methods of this kind, particularly for intricate details like the logo of a product, as shown in Fig.~\subref*{fig:d_b}, which poses significant risks in commercial applications. Later text-to-image methods in this field~\cite{li2023blip,chen2023anydoor}, which feature zero-shot subject-driven text-to-image generation, also suffer from poor subject fidelity.

Text-guided image inpainting methods~\cite{zhang2023adding, rombach2022high, DBLP:conf/cvpr/LiLZQWJ22} aim to fill missing regions within an image with the guidance of text prompts. They preserve the unmasked region precisely and fill the masked region to complete the image. Inspired by these methods, we consider a symmetrical task (\ie subject-preserved text-to-image generation) where the given subject is defined as the unmasked regions, and the large masked regions (background) are generated according to the text prompt. This task provides unique challenges to models' generative capacity and scene understanding ability, since the generated regions need to harmonize with the subject and be semantically reasonable. A significant shortcoming preventing current methods from being directly applied to the above task is their inability to determine the size and position of the subject automatically. Random placement often results in misplaced subjects, leading to the generation of unrealistic or unreasonable images, such as subjects floating in mid-air. Furthermore, these methods sometimes miss important scene objects when the text prompts are complex. Another issue arises from the fact that these approaches learn to fill a randomly masked region during training, such as brushes and squares~\cite{lugmayr2022repaint, DBLP:conf/cvpr/0001SMPNPOLFSB023}. But for our task, they need to fill the large background area with only the view of a complete subject. This discrepancy leads to degenerated images. How to address the above issues is still an open problem.

\begin{figure*}[t]
    \centering
    \includegraphics[width=\textwidth]{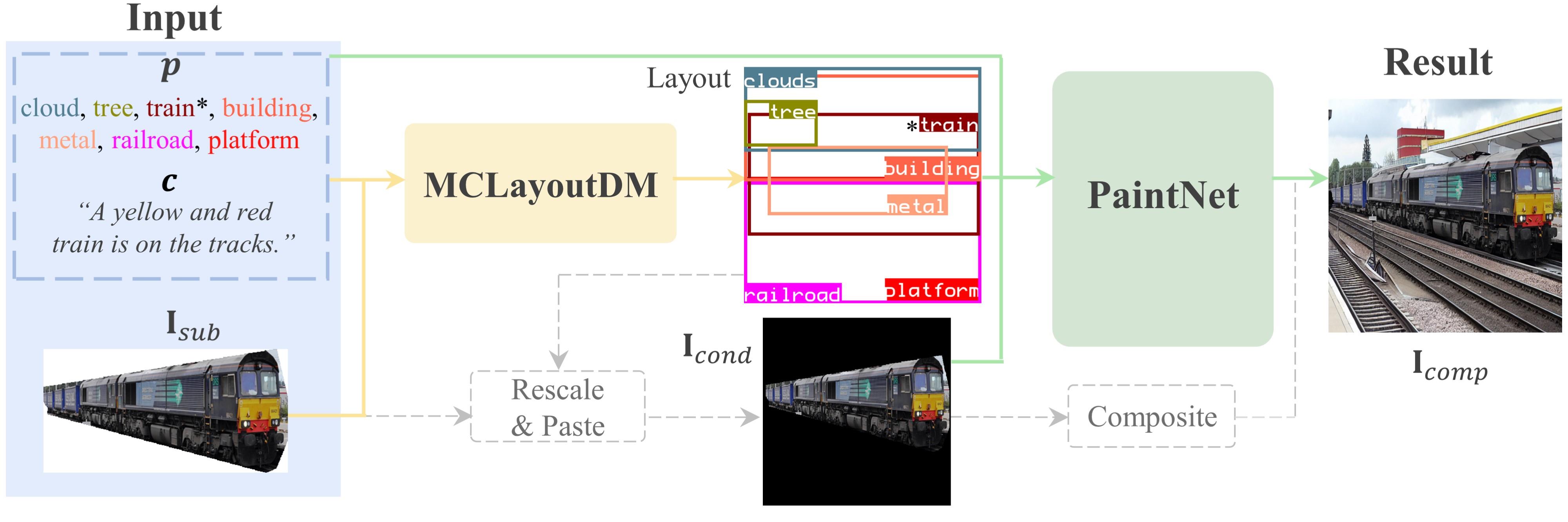}
    \caption{Overview of our proposed SceneBooth. It consists of a layout generation module, MCLayoutDM, and a background painting module, PaintNet. We use the ``*" symbol to mark the subject to preserve.}
    \label{fig:overview}
\end{figure*}

Inspired by the above insights, we propose a two-stage framework (i.e., SceneBooth) for subject-preserved text-to-image generation (Fig.~\ref{fig:overview}). Given as input a subject image, several object phrases and a scene caption, our framework generates high-quality images that contain the high-fidelity subject and align with the phrases and caption. The first stage aims to generate plausible scene layouts through a diffusion-based multimodal-conditioned layout generation module based on LayoutDM~\cite{chai2023layoutdm}. Specifically, we utilize the text-encoder and image-encoder in a pre-trained CLIP~\cite{radford2021learning} to extract textual and visual embeddings from multimodal inputs. These embeddings, containing rich contextual information, are then used as conditions by concatenation to generate high-quality scene layouts. The resulting layouts not only determine the position and size of the subject, but also provide an abstract and coarse position description about each object in the scene. The second stage aims to generate a background that harmonizes with the subject with the guidance of the caption, object phrases, and the layout generated in the first stage. We build a module named PaintNet based on pre-trained Latent Diffusion Model (LDM)~\cite{rombach2022high} and incorporate two kinds of adapters, Gated Self-Attention~\cite{li2023gligen} and ControlNet~\cite{zhang2023adding}, to simultaneously leverage the layout and visual conditional inputs. Unlike inpainting models, PaintNet is trained with instance masks. That is, PaintNet learns to generate a complete image when it has only the view of the subject. This training strategy encourages harmonized images in our setting, since the model learns that the unmasked regions contain a complete subject and tend to generate backgrounds harmonizing with the subject. Extensive experiments on COCO~\cite{cocodataset} dataset demonstrate the effectiveness of our approach.

Our contributions can be summarized as follows: 
{
\begin{itemize}
    \item We propose SceneBooth, a novel text-to-image framework that generates images with the given subject being faithfully preserved. Compared with existing methods, SceneBooth is of desired properties such as high-fidelity subject preservation, harmonious image generation, and strong scene coherence.
    \item We design a diffusion-based multimodal-conditioned layout generation module to generate reasonable scene layouts,
    and a text-to-image diffusion model based background generation module to generate a background harmonizing with the subject, with visual/textual and layout information guidance.
    \item Extensive experiments demonstrate that our framework outperforms existing methods in terms of subject preservation and visual perceptual quality.
\end{itemize}
}

\section{Related Work}

\subsection{Subject-driven Text-to-Image Generation}

Subject-driven text-to-image generation aims to generate images for a specific subject given several images of it and relevant text prompts. Most existing methods are based on Diffusion Models~\cite{ho2020denoising, sohl2015deep, song2020score, nichol2021improved}, which have demonstrated remarkable performance in the field of text-to-image generation~\cite{chai2023layoutdm,dhariwal2021diffusion,DBLP:conf/ijcai/HuangL0S00Z22}. Some methods~\cite{Ruiz2022DreamBoothFT, DBLP:conf/iclr/GalAAPBCC23, DBLP:journals/corr/abs-2305-03374, kumari2023multi} embed the given subject into the output domain of the model by finetuning on subject images. They need individual finetuning for each subject, and often fail to generate images that preserve the subjects accurately, especially in terms of the details. BLIP-Diffusion~\cite{li2023blip} and IntanceBooth~\cite{shi2023instantbooth} support zero-shot generation, but they also can only roughly preserve the style and appearance of the subject. Other works~\cite{wei2023elite, liu2023cones, voynov2023p, chen2024subject, alaluf2023neural} make progress in various aspects, but still leave much room for improvement in terms of subject fidelity. 

\subsection{Controlling Text-to-Image Diffusion Models}

The ability to customize or control large-scale text-to-image diffusion models for downstream tasks holds promising application value. To handle diverse control conditions, LDM~\cite{rombach2022high} trains task-specific models for each control condition, but this process is expensive. To address this, other methods~\cite{li2023gligen, zhang2023adding, DBLP:journals/corr/abs-2302-08453, DBLP:journals/corr/abs-2305-16322} adopt a more efficient way by adding a small number of task-specific parameters, known as adapters, to the pretrained base model and training only these newly added parameters. GLIGEN~\cite{li2023gligen} introduces Gated Self-Attention layers to the transformer blocks, enabling layout-guided text-to-image generation. ControlNet~\cite{zhang2023adding} maintains a trainable copy of Unet Encoder to produce conditioning features, achieving control with various spatially-aligned conditions. T2I-Adapter~\cite{DBLP:journals/corr/abs-2302-08453} employs a simple and lightweight adapter to achieve fine-grained control in the color and structure of the generated images.

\subsection{Scene Layout Generation}

Automatic layout generation for natural scenes has gained increasing attention. LayoutVAE~\cite{DBLP:conf/iccv/JyothiDHSM19} and LayoutGAN~\cite{DBLP:conf/iclr/LiYHZX19,DBLP:journals/tvcg/LiY0LWX21} are the first attempts to employ deep generative models to generate scene layouts. VTN~\cite{DBLP:conf/cvpr/ArroyoPT21} enhances diversity and quality by leveraging a self-attention based VAE. LayoutTransformer~\cite{DBLP:conf/iccv/GuptaLA0MS21, liang2023layout} and BO-GAN~\cite{DBLP:conf/mm/ChenMFH22} define layouts as discrete sequences and exploit the efficiency of transformers in generating structured sequences. LayoutDM~\cite{chai2023layoutdm} leverages the generation capabilities of diffusion models, thereby enhancing both quality and diversity. Most recently, some large language model based methods have also been explored~\cite{DBLP:conf/mm/QuW0NC23, DBLP:journals/corr/abs-2305-13655}.

\section{Our Method}
\label{sec:our_method}

\subsection{Problem Formulation}

We make a few assumptions about the inputs of our task: 
\textit{First}, the texts to describe each object in the image (including the subject) are given, which we refer to as object phrases. An object phrase can be a short descriptive sentence, such as \textit{``a blue shirt"}, or just a category label like \textit{``shirt"}. \textit{Second}, we assume each image contains exactly one subject we want to preserve. 
\textit{Third}, the caption gives an overall description of the entire scene, but without the necessity of mentioning all the objects in object phrases.

Given the image of a subject $\textbf{I}_{sub}$, a caption $\bm{c}$ to describe the scene, and several descriptive object phrases $\bm{p}=\{p_1, \cdots, p_N\}$ for each scene object, the goal of subject-preserved text-to-image generation is first to determine the position and scale of the subject and then generate a background that aligns with the caption and object phrases, and harmonizes with the subject.

\subsection{Overview}
Fig.~\ref{fig:overview} presents an overview of our two-stage subject-preserved text-to-image framework, SceneBooth. 
It consists of two main components: the MCLayoutDM module for scene layout generation with multi-modal conditional inputs, and the PaintNet module for background painting. Formally, our framework is described as follows:
\begin{align}
    \bm{l} &= \text{MCLayoutDM}(\mathbf{I}_{sub}, \bm{p}, \bm{c}) \\
    \textbf{I}_{cond} &= \textit{R\&P}(\textbf{I}_{sub}, \bm{l}) \\
    \mathbf{I}_{comp} &= m\odot \text{PaintNet}(\mathbf{I}_{cond}, \bm{p}, \bm{c}, \bm{l}) + (1 - m) \odot \textbf{I}_{cond}
\end{align}
where $\mathbf{I}_{sub}$, $\mathbf{I}_{cond}$ and $\mathbf{I}_{comp}$ represent the subject image, conditioning image and the completed image respectively. $\bm{c}$ is the caption describing the entire scene. $\bm{p}=[p_1, \cdots, p_i, \cdots, p_N]$ is a list of object phrases, such as $(``dog", ``grass", ``sky")$, indicating each object in the scene. And $\bm{l}=(l_1, \cdots,l_i, \cdots, l_N)$ is the generated layout of the scene, where $l_i=[x,y,w,h]$ corresponds to $p_i$, indicating the size and position of the $i$-th object in the scene. \textit{R\&P} is an operation that first \textit{rescales} the subject according to its bounding box in $\bm{l}$ and then \textit{pastes} it onto a blank canvas. 
$\odot$ is the element-wise product operation and $m$ is a binary mask corresponding to the subject region in the conditioning image, with a value of 0 for subject pixels and 1 for unknown background pixels.

Our framework adopt a coarse-to-fine approach to create images with specific subjects. First, MCLayoutDM generates a scene layout positioning the subject and background objects. Then, PaintNet produces a harmonious background for the subject based on this layout. We provide detailed descriptions of these two modules in the following sections.

\begin{figure*}[t!]
\captionsetup[subfloat]{font={scriptsize,rm},labelfont=scriptsize}
\centering
\subfloat[Layout Denoiser]{\includegraphics[width=0.46\linewidth]{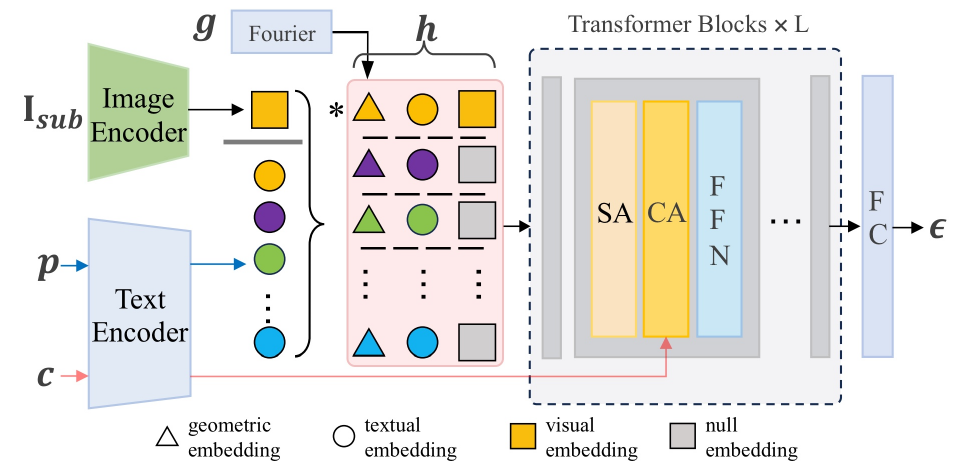}%
\label{fig:a}}
\hfil
\subfloat[Architecture of PaintNet]{\includegraphics[width=0.46\linewidth]{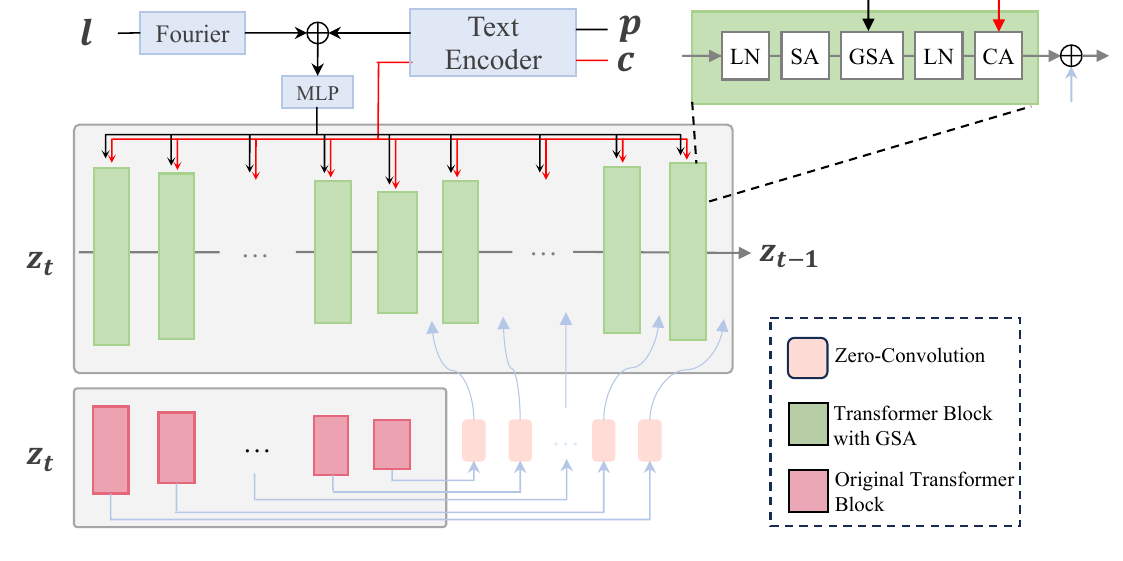}%
\label{fig:b}}
\caption{(a) Architecture of the layout denoiser in MCLayoutDM. Fourier, SA, CA, and FFN denote the fourier embedding layer, self-attention layer, cross-attention layer, and feed-forward network respectively. We use the ``*" symbol to mark the feature embeddings representing the subject. For simplicity, we omit the layer normalization and skip connections in the Transformer blocks, as well as the diffusion timestep input $t$. (b) Architecture of the PaintNet. LN and GSA denote layer normalization and Gated Self-Attention, respectively. (Best viewed in color.)}
\label{fig_sim}
\end{figure*}

\subsection{Multimodal-Conditioned LayoutDM}

Our multimodal-conditioned layout generation module MCLayoutDM is developed based on LayoutDM~\cite{chai2023layoutdm}. The inputs to MCLayoutDM are the caption $\bm{c}$, the subject image $\mathbf{I}_{sub}$, and several object phrases $\bm{p}$. Similar to LayoutDM, we design a Transformer-based~\cite{DBLP:conf/nips/VaswaniSPUJGKP17} layout denoiser and transform the layout generation into an iterative denoising process from pure Gaussian noise. 
The architecture of the multimodal-conditioned layout denoiser in MCLayoutDM is illustrated in Fig.~\subref*{fig:a}, which can be formalized as follows:
\begin{align}
    \bm{v}_{feat.} &= f_{vision}(\mathbf{I}_{sub}) \\
    \bm{p}_{feat.} &= [f_{text}(p_1), \cdots, f_{text}(p_N)] \\
    \bm{c}_{feat.} &= f_{text}(\bm{c}) \\ 
    \bm{g}_{feat.} &= \mathcal{F}(\bm{g}) \\
    [h_1, h_2, \cdots, h_N]&= \textit{Cat}(\bm{g}_{feat.}, \bm{p}_{feat.}, \textit{PAD}(\bm{v}_{feat.})) \\
    [h'_1, h'_2, \cdots, h'_N] &= \text{TB}( [h_1, h_2, \cdots, h_N]; \bm{c}_{feat.}) \\
    [\epsilon_1, \epsilon_2, \cdots, \epsilon_N] &= \text{FC}([h'_1, h'_2, \cdots, h'_N])
\end{align}
where $N$ is the number of objects in the scene. $\bm{g}=[g_1, g_2, \cdots, g_N]$ is the noised geometric parameters of the layout. $f_{text}$ and $f_{vision}$ are the text and image encoders in CLIP~\cite{radford2021learning}, respectively. $\mathcal{F}$ is the Fourier embedding~\cite{mildenhall2021nerf}. $\bm{v}_{feat.}$, $\bm{p}_{feat.}$, $\bm{c}_{feat.}$ and $\bm{g}_{feat.}$ are the encoded multimodal feature embeddings. \textit{Cat} and \textit{PAD} represent concatenate and padding operations respectively. TB indicates transformer block and FC represents fully connected layer.

\smallskip
\noindent \textbf{Feature Embedding.} 
We randomly scale the subject image $\mathbf{I}_{sub}$ within a 20\% range and paste it onto a fixed-sized (512×512 in our experiments) blank canvas before extracting visual features.
This data augmentation serves two purposes: First, it encourages the network to focus on the semantic and aspect ratio information of the subject image without leaking positional information. Second, it facilitates batch training.
Next, we adopt the ViT-based~\cite{dosovitskiy2020image} visual encoder from a pre-trained CLIP~\cite{radford2021learning} to obtain a feature vector $\bm{v}_{feat.}$ from the augmented subject image.
For the object phrases $\bm{p}$, we encode each $p_i$ with the text-encoder from the CLIP and construct a vector sequence with length $N$.
For the caption $\bm{c}$, we also obtain its text feature embedding $\bm{c}_{feat.}$ with CLIP text encoder. Benefiting from the vast concept knowledge in pre-trained vision/language models, the extracted multimodal feature embeddings provide rich information about the scene.

\smallskip
\noindent \textbf{Object Token Embedding.} We prepare object tokens that contain multimodal information for the transformer blocks. First, we employ the Fourier embedding to map the noised geometric parameters $\bm{g}$ to a higher-dimensional vector $\bm{g}_{feat.}$, enhancing the representation of high-frequency information~\cite{mildenhall2021nerf,rahaman2019spectral}. Next, we use a learnable null-vector to pad the visual embedding of the subject into a sequence of length $N$, since we can not obtain the visual information of the background objects. Finally, we concatenate these feature embeddings (geometric, textual, and visual) to construct the object tokens for each object in the scene.

\smallskip
\noindent \textbf{Transformer Blocks.}
Following~\cite{chai2023layoutdm}, we stack multiple transformer blocks to capture the relationships between scene objects from object tokens. We extend the original transformer blocks in~\cite{chai2023layoutdm} by adding a single cross-attention layer between the self-attention layer and the feed-forward network. This cross-attention layer allows the denoising process to be guided by the text feature embedding $\bm{c}_{feat.}$.

\subsection{PaintNet}
Given as inputs the conditional image $\textbf{I}_{cond}$, scene layout $\bm{l}$, caption $\bm{c}$ and object phrases $\bm{p}$, our background painting module, PaintNet, generates high-quality images with backgrounds that harmoniously integrate with the subjects. We build PaintNet based on LDM~\cite{rombach2022high}, and incorporate two types of adapters, namely Gated Self-Attention~\cite{li2023gligen} and ControlNet~\cite{zhang2023adding}, to leverage different conditional inputs. The architecture of PaintNet is illustrated in Fig.~\subref*{fig:b}. 

\smallskip
\noindent \textbf{Gated Self-Attention.} We inject the layout information into LDM by utilizing the Gated Self-Attention~\cite{li2023gligen} layer. It performs a special attention operation on the concatenation of visual tokens and specifically-designed grounding tokens.
Following~\cite{li2023gligen}, we add gated self-attention layers between the self-attention and cross-attention layers. The construction of grounding tokens and the computation of gated self-attention can be described as follows:
\begin{align}
    d_i&=\text{MLP}(f_{text}(p_i), \text{Fourier}(l_i)) \\
    \bm{v} &= \bm{v} + \beta \cdot \text{tanh}(\gamma)\cdot \text{TS}(\text{SelfAttn}([\bm{v}, \bm{d}]))
\end{align}
where $p_i$ and $l_i$ are the object phrase and geometric parameters of $i$-th object in the scene, $\bm{v}=[v_1, \cdots, v_M]$ and $\bm{d}=[d_1, \cdots, d_N]$ are the visual feature tokens and grounding tokens respectively. $\text{TS}(\cdot)$ is a token selection operation that considers visual tokens only, and $\gamma$ is a learnable scalar which is initialized as 0. Following~\cite{li2023gligen}, we set $\beta$ as 1.

\smallskip
\noindent \textbf{ControlNet.} We inject the visual information of the subject into LDM through ControlNet~\cite{zhang2023adding}. Specifically, we obtain conditioning features from $\textbf{I}_{cond}$ using a trainable copy of the Unet Encoder in LDM and add them back to the Unet through zero-convolution layers. The key is constructing appropriate conditioning images $\textbf{I}_{cond}$. 
During training, we randomly extract a subject/instance from each image using ground-truth segmentation annotations. We then normalize the pixel values of the subject to $[0, 1]$ and assign other pixel values as $-1$ to make up the conditioning images. Note here that, the main difference in our training, as compared to the ControlNet-inpaint~\cite{zhang2023adding} method, lies in our use of ``instance masks" rather than random masks to construct the conditioning image. During inference, the conditioning images are constructed by rescaling and pasting the given subject into its corresponding bounding box and performing the same value mapping operation as training time.

\subsection{Training Objective}
Both MCLayoutDM and PaintNet are based on $\bm{\epsilon}$-prediction~\cite{ho2020denoising, rombach2022high} diffusion models and are trained using a denoising objective as follows:
\begin{equation}
\mathop{\mathrm{min}}\limits_{\bm{\theta}'}\mathcal{L}=\mathbb{E}_{\bm{z},\bm{\epsilon}\sim\mathcal{N}(\mathbf{0},\mathbf{I}),t}\left[\|\bm{\epsilon}-\bm{\epsilon}_{\{\bm{\theta},\bm{\theta}'\}}(\bm{z}_t, t, \bm{y})\|^2\right]
\end{equation}
where $t$ is uniformly sampled from time steps $\{1,\cdots, T\}$, $\bm{z}_t$ is the noised variant of input $\bm{z}$ at timestep $t$, $\bm{y}$ is the conditional input, and $\bm{\epsilon}_{\{\bm{\theta}, \bm{\theta}'\}}(\cdot)$ is the neural denoiser with frozen parameters $\theta$ and trainable parameters $\theta'$. 

For MCLayoutDM, $\bm{z}_t$ is the noised layout geometric parameters and $\bm{y}=(\mathbf{I}_{sub},\bm{p},\bm{c})$. We freeze the image and text encoders and train the transformer blocks and fully connected layers. For PaintNet, $\bm{z}_t$ is the noised latent vector and $\bm{y}=(\mathbf{I}_{cond},\bm{p}, \bm{c},\bm{l})$. We freeze the original Unet, and train the Gated Self-attention layers, zero-convolution layers and the trainable copy of Unet encoder.

\begin{table*}[t!]
\renewcommand\arraystretch{1.1}
\setlength{\tabcolsep}{11pt}
\centering
\caption{Comparison with existing text-to-image inpainting methods on COCO.  ``*" denotes COCO finetuned baselines}

\begin{tabular}{l|c|c|c|c|cccc}
\hline
Method                  & FID↓           & CLIP-T↑         & CLIP-I↑         & DINO↑           & \multicolumn{1}{c|}{$\textbf{P}_{quality}$}        & \multicolumn{1}{c|}{$\textbf{P}_{fidelity}$}       & \multicolumn{1}{c|}{$\textbf{P}_{{obj}_f}$}           & $\textbf{P}_{{prompt}_f}$        \\ \hline
ControlNet-inpaint       & 31.38          & 0.3094          & 0.7334          & 0.6469          & \multicolumn{4}{c}{\multirow{2}{*}{-}}                                                                                               \\ \cline{1-5}
StableDiffusion-inpaint  & 24.15          & 0.3116          & 0.7558          & 0.6950          & \multicolumn{4}{c}{}                                                                                                                 \\ \hline
ControlNet-inpaint*      & 29.24          & 0.3106          & 0.7482          & 0.6600          & \multicolumn{1}{c|}{6.4\%}           & \multicolumn{1}{c|}{8.6\%}           & \multicolumn{1}{c|}{16.7\%}          & 18.3\%          \\ \hline
StableDiffusion-inpaint* & 23.33          & \textbf{0.3132} & 0.7855          & 0.7062          & \multicolumn{1}{c|}{42.2\%}          & \multicolumn{1}{c|}{42.0\%}          & \multicolumn{1}{c|}{18.9\%}          &   30.0\%        \\ \hline
\textbf{SceneBooth(Ours)}         & \textbf{20.89} & 0.3112          & \textbf{0.8023} & \textbf{0.7501} & \multicolumn{1}{c|}{\textbf{51.4\%}} & \multicolumn{1}{c|}{\textbf{49.4\%}} & \multicolumn{1}{c|}{\textbf{64.4\%}} & \textbf{51.7\%} \\ \hline
\end{tabular}

\label{tab:layout-agnostic}
\end{table*}

\section{Experiments}

\subsection{Experimental Setup}

\noindent \textbf{Dataset.} We evaluate our framework on \lzh{COCO2017~\cite{cocodataset}, a large-scale multimodal dataset with detailed annotations for object detection, segmentation, and captioning tasks.}
We select samples containing annotations for semantic segmentation, layout, and caption from COCO2017, which contain objects covering 80 categories of things and 91 categories of stuff. To eliminate unnecessarily complicated or too simple cases, we filter out samples with more than 8 or less than 3 objects. We use 95\% of the official training split for training, the rest for validation, and the official validation split for testing. The dataset consists of 65k/3.4k/2.8k for training/validation/testing.

\smallskip
\noindent \textbf{Evaluation Metrics.} We evaluate the quality of the generated images with FID~\cite{DBLP:conf/nips/HeuselRUNH17}, CLIP-T~\cite{Ruiz2022DreamBoothFT}, CLIP-I~\cite{DBLP:conf/iclr/GalAAPBCC23} and DINO~\cite{Ruiz2022DreamBoothFT} metrics. 
\textbf{FID}~\cite{DBLP:conf/nips/HeuselRUNH17} calculates the distribution distance between real and generated samples.
\textbf{CLIP-T}~\cite{Ruiz2022DreamBoothFT} measures the alignment between the prompt and the generated image by computing CLIPScore\cite{hessel2021clipscore} between them. \textbf{DINO}~\cite{Ruiz2022DreamBoothFT} and \textbf{CLIP-I}~\cite{DBLP:conf/iclr/GalAAPBCC23} measure the subject fidelity. We further employ additional metrics to evaluate the alignment between images and layouts, and the quality of the generated layouts.
\textbf{YOLO Score}~\cite{li2021image, li2023gligen} evaluates whether the layout of the generated image is consistent with the input layout. \textbf{Max. IoU}~\cite{DBLP:conf/mm/KikuchiSOY21} measures the similarity between the set of generated layouts and the ground-truth set. It computes the highest layout IoU under optimal matching. In the original implementation of Max. IoU, only one layout is generated per input sample, and a match occurs when the input object phrases are the same. We make an extensions to this metric by generating $k$ layouts per input sample, referred to as Max. IoU @ k. Please refer to supplementary materials for more details.

\textbf{Human Evaluations} are designed for further evaluations. We randomly sample 200 metadata records from the test set and use these metadata to generate 200 images using different methods. For each record, 5 annotators are asked to pick the generated image with the best overall quality, the best subject fidelity, the closest match to the given object phrases, and the best alignment with the caption. For example, the question concerning overall quality is ``Please examine each image carefully and select the one you believe has the highest overall quality. Consider factors such as clarity, realism, and composition.". To prevent potential bias, the subject is shown to annotators only when evaluating subject fidelity. The human evaluations are assigned to a third-party annotation company staffed with experienced annotators in the computer vision field. The percentages of images being chosen as the best ones are denoted as $\textbf{P}_{quality}$, $\textbf{P}_{fidelity}$, $\textbf{P}_{obj_f}$ and $\textbf{P}_{prompt_f}$ for each method.

\smallskip
\noindent \textbf{Implementing Details.}
Our framework is implemented with PyTorch~\cite{Paszke_PyTorch_An_Imperative_2019}. We train MCLayoutDM using Adam optimizer~\cite{kingma2014adam} with a learning rate of 1e-5 and the batch size is set to 64. The training contains 400k iterations, taking about 30 hours on a single NVIDIA V100 GPU. When training PaintNet, we use Adam optimizer with a learning rate 5e-5 and the batch size is set to 8. The model is trained with 102k iterations, taking about 110 hours on 8 NVIDIA A100 GPUs. We use pre-trained CLIP~\cite{radford2021learning} to initialize the text/image encoder weights in MCLayoutDM and PaintNet. When training PaintNet, we initialize the weights of Unet with those from Stable Diffusion v1.5~\cite{rombach2022high}. Our codes and weights will be released after this paper is published.

\subsection{Comparison with existing methods}

\begin{figure*}[t!]
    \centering
    \includegraphics[width=\linewidth]{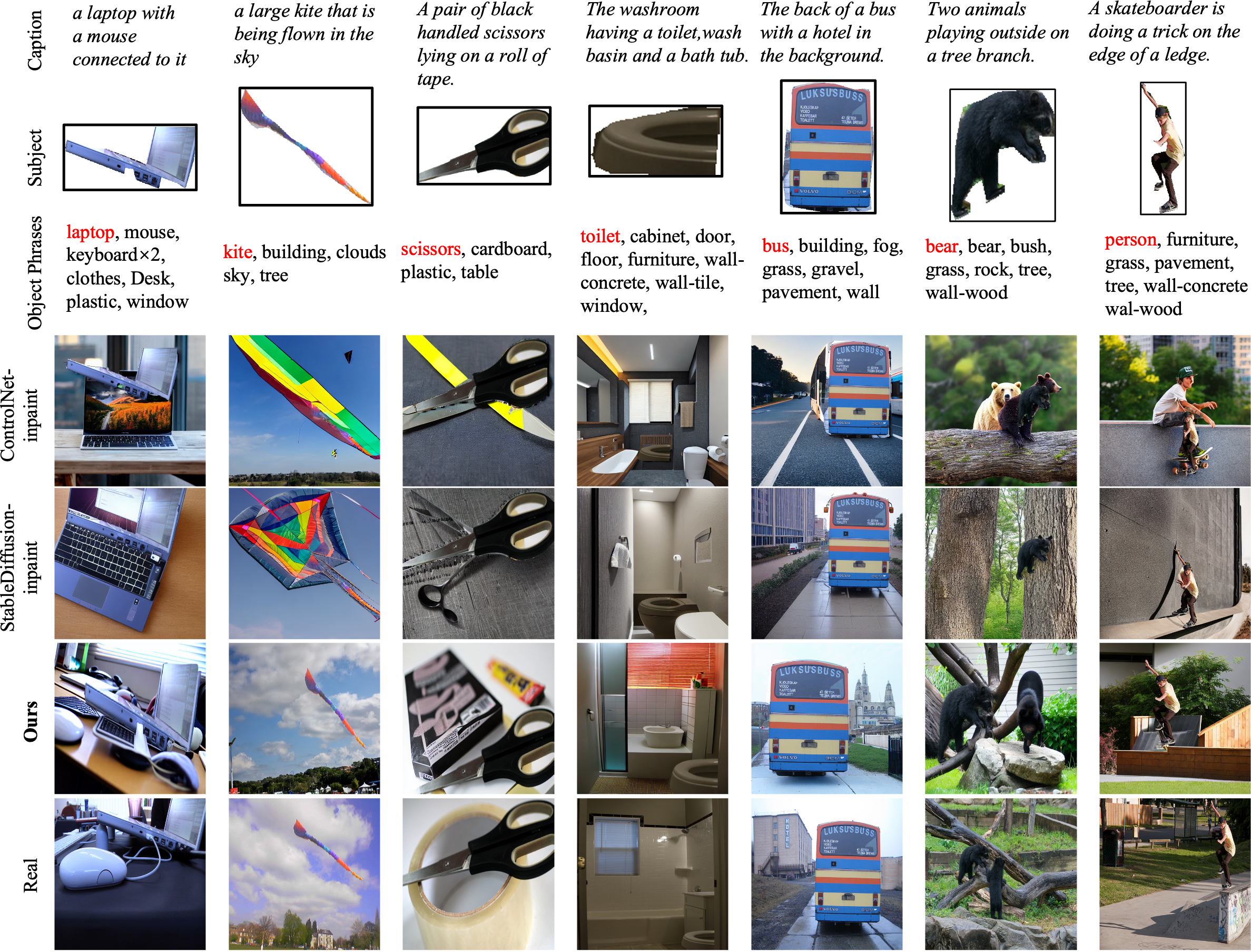}
    \caption{Qualitative comparison with existing methods on COCO dataset. The subject in object phrases is highlighted in red.}
    \label{fig:two_stage}
\end{figure*}

Our task setting differs notably from prior works, with image inpainting being the closest related task. To showcase the strengths of our framework, we compare it with two inpainting methods built on large-scale text-to-image models: StableDiffusion-inpaint~\cite{rombach2022high} and ControlNet-inpaint~\cite{zhang2023adding}.
\lzh{Both methods are finetuned on COCO~\cite{cocodataset} for fair comparison.} 
Note here that: (1) Our framework receives additional scene information from the object phrases compared to comparison methods. To narrow the potential gap, we follow~\cite{DBLP:journals/corr/abs-2309-14303} and append the caption with the object phrases to create new text prompts for the comparison methods. 
(2) We generate images with the subject placed at random positions and with a 20\% variation in scale when evaluating comparison methods, as they are unable to determine the exact position or scale of the subject in the output image.

\noindent \textbf{Quantitative Evaluation.} 
The quantitative comparison results are shown in Table~\ref{tab:layout-agnostic}. 
We can see that: (1) Our method significantly outperforms the other two methods in terms of the FID metric, demonstrating that the images generated by our method are closer to real images and of overall higher quality and diversity, since FID captures both aspects.
(2) Our method significantly outperforms the comparison methods regarding the CLIP-I and DINO metrics, indicating better performance in subject fidelity. This improvement is attributed to our framework's subject-preserved generative process and the deep understanding and utilization of layout information along with subject visual information. (3) There is little difference among the three methods on CLIP-T, indicating no significant gap in the image-text semantic alignment in the feature space. (4) In human evaluations, our approach achieves highest scores on all four metrics. Specifically, our method demonstrates evident improvement in terms of $\textbf{P}_{obj_{f}}$ and $\textbf{P}_{prompt_{f}}$. This is because our method offers layout-level guidance over objects in an image, thereby enhancing counting and positional relationships, which is difficult to capture with CLIP-T metric. However, the superiority on $\textbf{P}_{quality}$ and $\textbf{P}_{fidelity}$ is comparatively modest. This is because our layout generation module sometimes produces less plausible layouts, resulting in reduced quality of results. In ablation studies, we observe a significant metric improvement when ground-truth layouts are given.

\noindent \textbf{Qualitative Evaluation.} 
We qualitatively compare the generation performance of our SceneBooth with StableDiffusion-inpaint and ControlNet-inpaint.
The results are displayed in Fig.~\ref{fig:two_stage}. We can see that: \lzh{both other methods generate images where the subject and background are not well blended, seems like a direct paste-on (column 1,3,6,7). In contrast, our approach can generate more natural images where the subject is seamlessly integrated into the scene. 
Results generated by our method also do not have obvious artifacts around the subjects like other methods. This demonstrates our framework can effectively learn how to integrate the subject into the scene harmoniously. 
}

\begin{table*}[t!]
\renewcommand\arraystretch{1.1}
\setlength{\tabcolsep}{11pt}
\centering
\caption{Ablation study on the effectiveness of ControlNet
}

\begin{tabular}{l|c|c|c|c|c|c|c}
\hline
Method                  & FID↓           & CLIP-T↑         & YOLO Score(AP/AP50/AP75)↑  & CLIP-I↑         & DINO↑           & $\textbf{P}_{quality}$        & $\textbf{P}_{fidelity}$       \\ \hline
GLIGEN-repaint          & 23.72          & 0.3084          & 0.286 / 0.417 / 0.323          & 0.7763          & 0.7153          & 7.7\%           & 5.7\%           \\ \hline
GLIGEN-inpaint          & 22.87          & 0.3099          & 0.251 / 0.398 / 0.274          & 0.7873          & 0.7326          & 11.1\%          & 11.0\%          \\ \hline
\textbf{PaintNet(Ours)} & \textbf{18.58} & \textbf{0.3130} & \textbf{0.292 / 0.419 / 0.332} & \textbf{0.8420} & \textbf{0.8094} & \textbf{81.2\%} & \textbf{83.3\%} \\ \hline
\end{tabular}

\label{tab:layout-aware}
\end{table*}

\begin{table}[t]
\caption{Ablation study on the effectiveness of visual embedding used in MCLayoutDM.
}
\renewcommand\arraystretch{1.1}
\setlength{\tabcolsep}{10pt}
\label{tab:layout_compare}
\centering

\begin{tabular}{l|c|c}
\hline
Method            & Max. IoU @(1/3/5)↑                                                                & FID↓                 \\ \hline
VTN~\cite{DBLP:conf/cvpr/ArroyoPT21}         & 0.295/0.357/0.387                                          & 24.29          \\ \hline
GAN++~\cite{DBLP:conf/mm/KikuchiSOY21} & 0.312/0.344/0.375  & 24.03 \\ \hline
BO-GAN~\cite{DBLP:conf/mm/ChenMFH22}      & 0.347/0.367/0.377                                                               & /                    \\ \hline
LayoutDM~\cite{chai2023layoutdm} & 0.377/0.383/0.410  & 21.55 \\ \hline
\textbf{MCLayoutDM(Ours)}       & \textbf{0.383}/\textbf{0.389}/\textbf{0.413}               & \textbf{20.89}                \\ \hline
\end{tabular}
\end{table}

\subsection{Ablation Study}

\smallskip
\noindent \textbf{Effectiveness of ControlNet.} ControlNet is one of the critical components in PaintNet. It receives the subject image input and injects the extracted visual information into the Unet structure. Removing ControlNet directly will change the input of the module, and cause the PaintNet to descend into an existing layout-guided text-to-image method: GLIGEN~\cite{li2023gligen}. To align the inputs and independently evaluate the effect of ControlNet, we compare the full PaintNet with two inpainting methods based on GLIGEN.
We refer to the inpainting method using the masked denoising strategy in~\cite{lugmayr2022repaint} as GLIGEN-repaint and the one with 5 additional Unet input channels~\cite{rombach2022high,li2023gligen}, specifically finetuned for the inpainting task, as GLIGEN-inpaint. Note here that we do not compare with layout-to-image methods such as LostGAN~\cite{Sun2019ImageSF,Sun2020LearningLA}, OC-GAN~\cite{Sylvain2020ObjectCentricIG} and LayoutDiffusion~\cite{Zheng2023LayoutDiffusionCD}, because they do not receive text prompt input and work at different resolutions from ours (64/128/256 vs 512). Ground-truth layouts are given as input to eliminate the impact of random layouts on performance evaluation.

Table~\ref{tab:layout-aware} reports the quantitative comparison results.
We can observe that: (1) Our PaintNet significantly outperforms GLIGEN-repaint and GLIGEN-inpaint on FID, which indicates that the images generated by PaintNet are more similar to the real images. This demonstrates the effectiveness of ControlNet in retaining the subject's appearance to generate realistic and harmonious images. (2) Results generated by PaintNet report an enhanced alignment with the layouts, as indicated by the higher Yolo score achieved. This is likely due to the ``instance mask" strategy we employ during training the ControlNet part, which effectively avoid redundant flaws around the subjects. (3) Regarding the CLIP-I and DINO metrics, PaintNet also showcases superior performance. 
This emphasizes ControlNet's superiority in better preserving the subjects' appearance compared to the other two methods.
(4) In human evaluations, PaintNet outperforms counterparts with a winning rate exceeding 80\% for both image quality ($\text{P}_{quality}$) and subject fidelity ($\text{P}_{fidelity}$) assessments. This further demonstrates the effectiveness of ControlNet.

\begin{figure*}[t!]
    \centering
    \includegraphics[width=\linewidth]{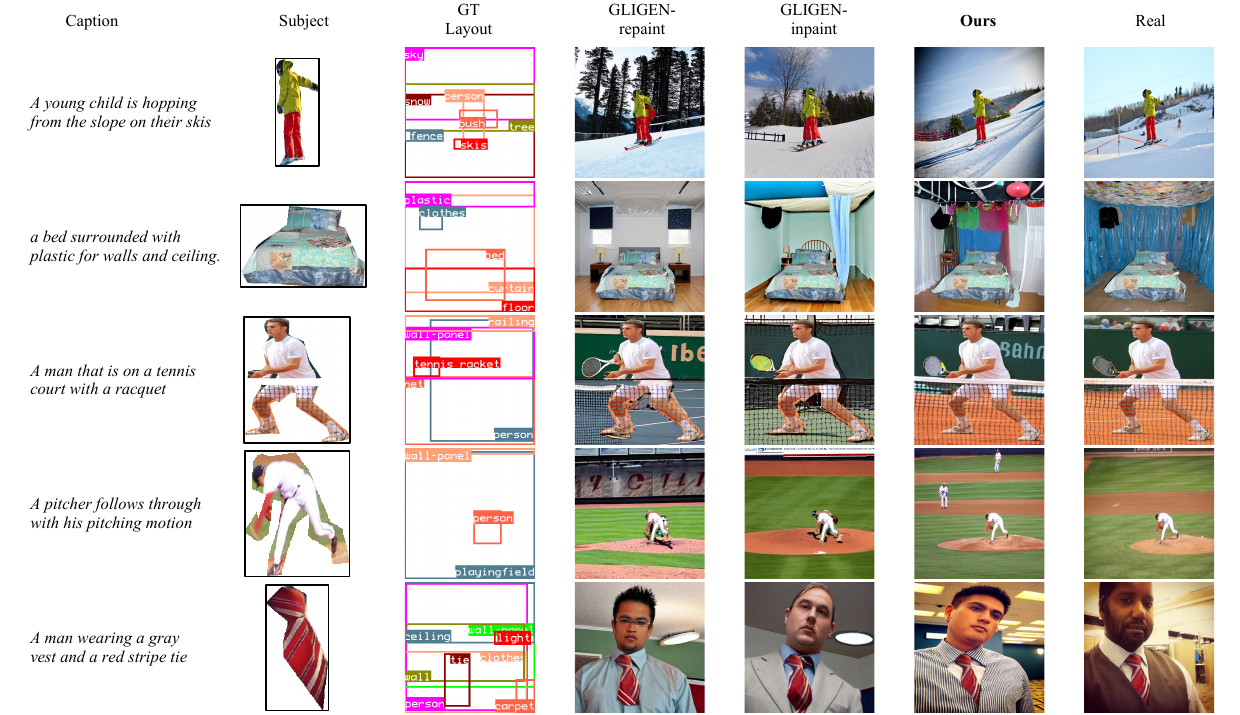}
    \caption{Ablation study on the effectiveness of ControlNet. We qualitatively compare PaintNet with GLIGEN-repaint and GLIGEN-inpaint on test dataset. Ground-truth layouts are used as input.}
    \label{fig:layout-aware-png}
\end{figure*}

Fig.~\ref{fig:layout-aware-png} displays the qualitative comparison results. One can observe that: Due to the introduction of ControlNet, our PaintNet can draw the background to match the context of the foreground subject, ensuring a harmonious and seamless blend. For example, in the third row, our PaintNet draws the tennis player on a clay court, in contrast to GLIGEN-repaint and GLIGEN-inpaint, which place the same player on a blue or green court, resulting in images that appear much more unnatural. In the first and the fourth rows, GLIGEN-repaint and GLIGEN-inpaint generate images with human subjects surrounded by evident artifacts, whereas PaintNet creates images with tidy edges where the subject and background are seamlessly integrated.

\begin{figure*}[t!]
    \centering
    \includegraphics[width=\linewidth]{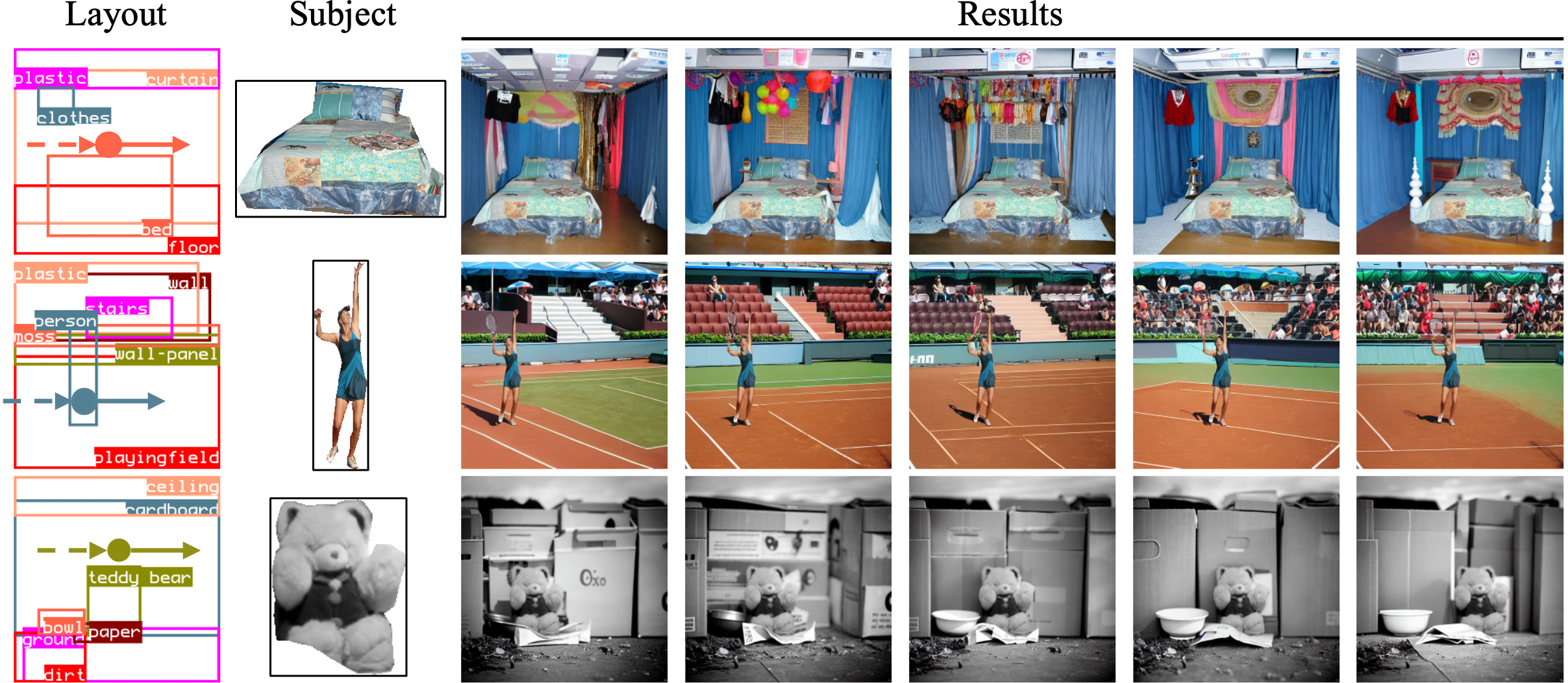}
    \caption{Qualitative results of subject translation in the scene layout. The direction of translation is indicated using dots and arrows. Row 1-3 have captions: ``a bed surrounded with plastic for walls and ceiling.", ``a girl getting ready to serve on the tennis court." and ``A teddy bear sitting on the ground in front of many boxes." respectively.}
    \label{fig:drag}
\end{figure*}

\smallskip
\noindent \textbf{Effectiveness of Subject Visual Embedding.}
Built on LayoutDM, our MCLayoutDM module introduces multimodal embeddings to facilitate layout generation under the guidance of both textual and visual inputs. 
Since the effectiveness of text embedding used in MCLayoutDM has been proved in former works~\cite{DBLP:conf/mm/ChenMFH22, liang2023layout}, we independently evaluate the effectiveness of the subject visual embedding. Besides LayoutDM~\cite{chai2023layoutdm}, we also select other layout generation method which do not receive visual input, such as BO-GAN~\cite{DBLP:conf/mm/ChenMFH22}, LayoutGAN++~\cite{DBLP:conf/mm/KikuchiSOY21}, and VTN~\cite{DBLP:conf/cvpr/ArroyoPT21}, as our baselines.
 VTN, LayoutGAN++ and LayoutDM do not receive text prompts as input, so we add cross-attention layers with text embedding to enable them to be guided by the text prompts. \lzh{To assess the reasonability of layouts, we employ PaintNet to generate images using the layouts from each method and calculate the FID score.} \lzh{Table~\ref{tab:layout_compare} presents the quantitative evaluation results, our approach outperforms other methods by a considerable margin on both metrics, showing that the proposed MCLayoutDM can better capture and model the relationship of different objects and generate more plausible layouts. Our model performs better than LayoutDM, indicating the subjects' visual feature facilitates the model to generate more reasonable layouts. We do not provide the FID values of BO-GAN, because it autoregressively predicts discrete layout sequences, which sometimes leads to missing or wrong object categories in final layouts. This makes fair comparison difficult.
}

\smallskip
\noindent \textbf{Different Attention Types.} 
The effect of different attention types is evaluated by changing the attention layers in PaintNet which process the grounding tokens. We generate images using ground-truth layouts and evaluate the results. Table~\ref{tab:attention_type} shows that Gated Self-Attention performs better, achieving comparable performance to Gated Cross-Attention on the CLIP-T, and significantly outperforming Gated Cross-Attention on the other three metrics.

\smallskip
\noindent \textbf{Different Mask Strategies.} The effect of different mask strategies is evaluated. As shown in Table~\ref{tab:mask_strategy}, using the ``instance mask" strategy obtains a higher performance than the ``random mask" strategy. ``instance mask" performs comparably to the original ``random mask" strategy on the CLIP-T metric, and significantly improves the performance on the FID/CLIP-I/DINO metrics.

\begin{table}[t]
\caption{Ablation study on attention type. 
``GCA" and ``GSA" means ``Gated Cross-Attention" and ``Gated Self-Attention"}
\label{tab:attention_type}
\centering
\renewcommand\arraystretch{1.1}
\setlength{\tabcolsep}{12pt}
\begin{tabular}{l|c|c|c|c}
\hline
    & FID↓           & CLIP-T↑         & CLIP-I↑         & DINO↑           \\ \hline
GCA & 21.05          & \textbf{0.3141} & 0.7922          & 0.7366          \\ \hline
\textbf{GSA} & \textbf{18.58} & 0.3130          & \textbf{0.8420} & \textbf{0.8094} \\ \hline
\end{tabular}
\end{table}

\begin{table}[t]
    \centering
    \caption{Ablation study on mask strategy. 
    ``random" and ``instance" denotes ``random mask" and ``instance mask"
}
\label{tab:mask_strategy}
\centering
\renewcommand\arraystretch{1.1}
\setlength{\tabcolsep}{11pt}
\begin{tabular}{l|c|c|c|c}
\hline
              & FID↓           & CLIP-T↑         & CLIP-I↑         & DINO↑           \\ \hline
random   & 21.36          & \textbf{0.3113} & 0.7854          & 0.7226          \\ \hline
\textbf{instance} & \textbf{20.89} & 0.3112          & \textbf{0.8023} & \textbf{0.7501} \\ \hline
\end{tabular}
\end{table}

\smallskip
\subsection{Extended Tasks}
\noindent \textbf{``Drag" your subject.}
In SceneBooth, the scene layout is generated by MCLayoutDM, and one boundingbox within it represents the subject that we wish to preserve. Taking inspiration from~\cite{pan2023drag}, we achieve local control over the position of the subject by ``dragging" its boundingbox. In this way, we can manipulate the position of the subject while perserving its appearance with high fidelity. The qualitative results are displayed in Fig.~\ref{fig:drag}.

\begin{figure*}[t!]
    \centering
    \includegraphics[width=0.97\linewidth]{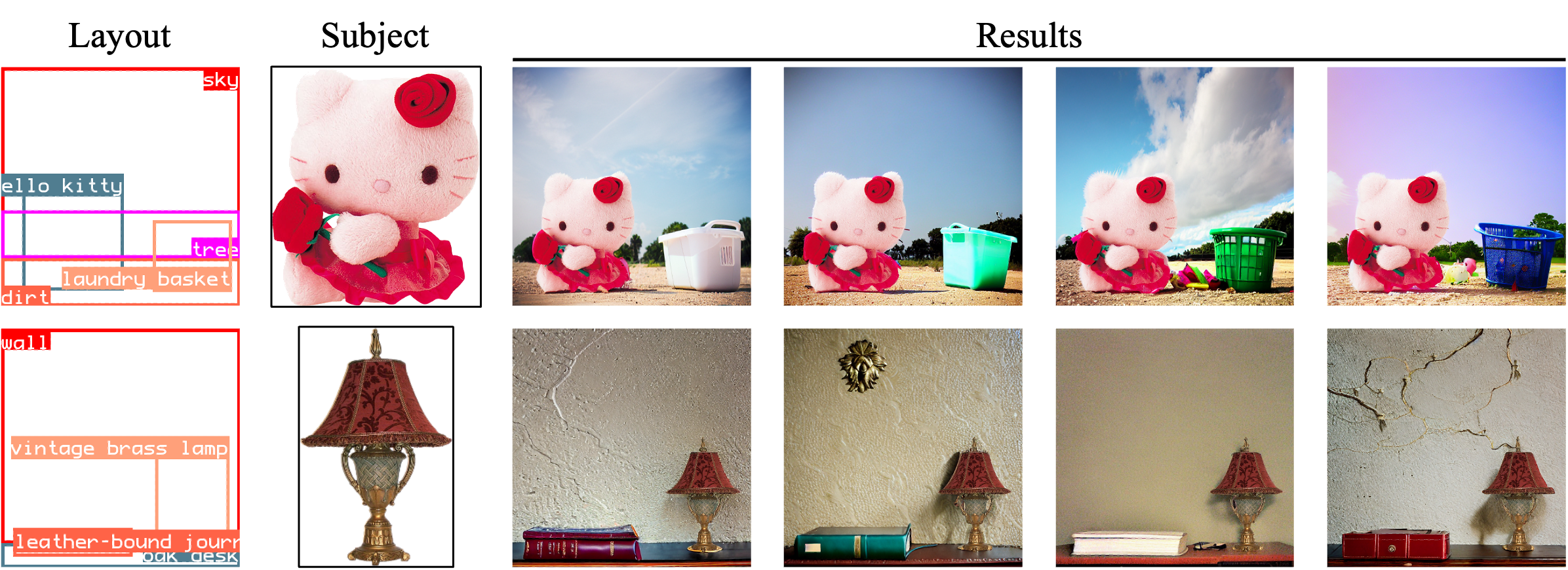}
    \caption{Generation under open-world setting. Row 1-2 have captions: ``A \textit{hello kitty} is sitting on the ground besides a \textit{laundry basket}" and ``A \textit{leather-bound journal} is resting on an oak desk next to a \textit{vintage brass lamp}."}
    \label{fig:open_world}
\end{figure*}

\smallskip
\noindent \textbf{Open-world generation.} With the knowledge from large-scale pre-trained text-to-image model in the open world (pre-trained Latent Diffusion Model in our paper), SceneBooth is able to generate objects that it has not seen in the training dataset (COCO dataset). For instance, Hello Kitty dolls and classic lamps. Figure 5.7 shows two examples of generation in open-world scenarios. From the generated results, We can observe that the method proposed in this paper can generate high-quality and personalized images with preserved target appearances in such open-world scenarios.

\begin{figure}
    \centering
    \includegraphics[width=\linewidth]{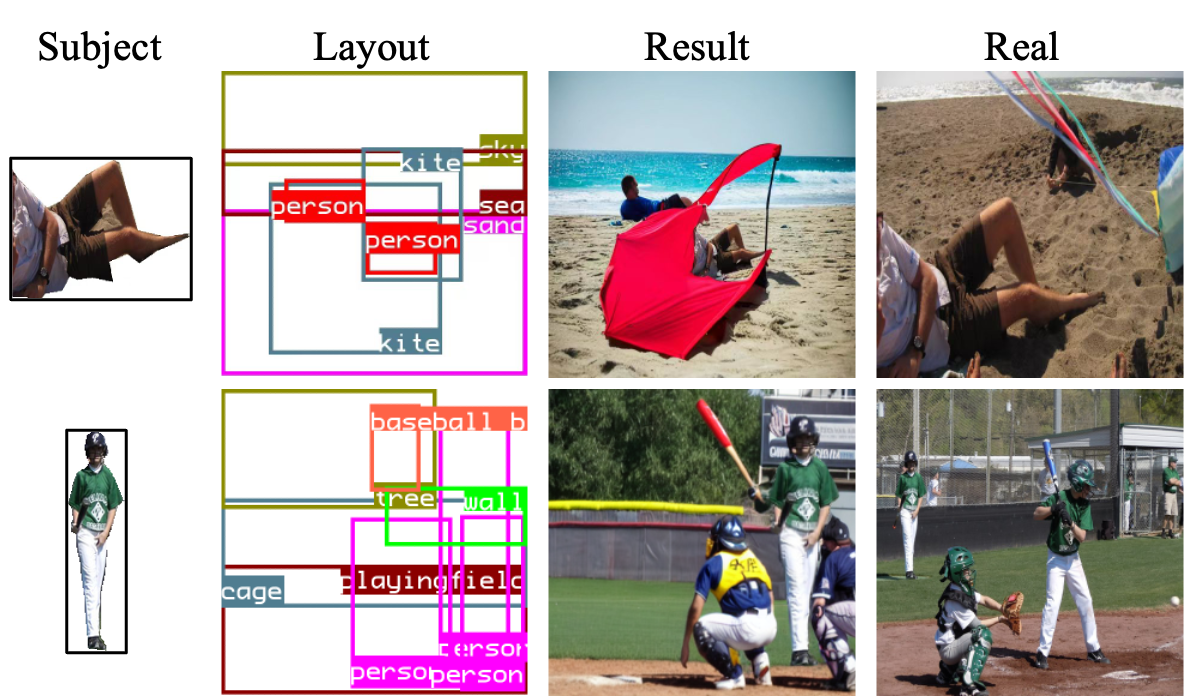}
    \caption{Two problematic cases. Row 1-2 have captions: ``A man in shorts is laying on the beach" and ``A baseball player on home plate swinging a bat". Real images are shown for reference.}
    \label{fig:failure}
\end{figure}

\section{Conclusion} 
This paper proposes a diffusion-based framework SceneBooth to address subject-preserved text-to-image generation. We introduce a multimodal-conditioned layout generation module to generate high-quality scene layouts which determines the position of the subject and other scene objects. Then, we employ a diffusion-based background generation module, which incorporates two kinds of adapters, to generate a harmonious background for the given subject with the guidance of the texts and layout. Quantitative and qualitative results demonstrate the impressive performance of our framework in subject fidelity and perceptual quality.

\noindent \textbf{Limitations.} Our background painting module sometimes struggles with partially occluded subject input (e.g., the first row in Fig.~\ref{fig:failure}), potentially due to the conflict between the relations of boundingboxes and the subject's pose. Another issue, as shown in the second row of Fig.~\ref{fig:failure}, is that generated layouts sometimes become irrational when there are too many objects, owing to insufficient training data for object-heavy scenarios.

\noindent \textbf{Future Directions.} Although our method shows plausible results in subject-preserved text-to-image generation in comparison to existing methods, it still has limitations. \lzh{First, our method can not handle situations where multiple subjects need to be preserved.  
Second, we do not impose strict constraints on the aspect ratio of the subjects during training MCLayoutDM, potentially leading to minor variations in practice. We leave the solutions to the above problems for future work.}


\bibliographystyle{IEEEtran}
\bibliography{trans}

\end{document}